\documentclass[runningheads]{llncs}
\usepackage[T1]{fontenc}
\usepackage{graphicx}
\usepackage{amsmath}
\usepackage{amsfonts}
\usepackage{bm}
\usepackage{listings}
\usepackage{xcolor}
\usepackage{subcaption}
\usepackage{stmaryrd}
\usepackage{booktabs}
\usepackage{hyperref}
\usepackage{float}

\definecolor{codegreen}{rgb}{0,0.6,0}
\definecolor{codegray}{rgb}{0.5,0.5,0.5}
\definecolor{codepurple}{rgb}{0.58,0,0.82}
\definecolor{backcolour}{rgb}{0.95,0.95,0.92}

\lstdefinestyle{mystyle}{
    backgroundcolor=\color{backcolour},   
    commentstyle=\color{codegreen},
    keywordstyle=\color{magenta},
    numberstyle=\tiny\color{codegray},
    stringstyle=\color{codepurple},
    basicstyle=\ttfamily\footnotesize,
    breakatwhitespace=false,         
    breaklines=true,                 
    captionpos=b,                    
    keepspaces=true,                 
    numbers=left,                    
    numbersep=5pt,                  
    showspaces=false,                
    showstringspaces=false,
    showtabs=false,                  
    tabsize=2
}

\begin{document}
\title{Guiding Registration with Emergent Similarity from Pre-Trained Diffusion Models}
\titlerunning{Guiding Registration with Pre-trained Diffusion Models}
\author{Nurislam Tursynbek\inst{1}, Hastings Greer\inst{1}, Ba\c{s}ar Demir\inst{1}, Marc Niethammer\inst{2}}
\authorrunning{N. Tursynbek et al.}
%\institute{Princeton University, Princeton NJ 08544, USA \and
%Springer Heidelberg, Tiergartenstr. 17, 69121 Heidelberg, Germany
\institute{\textsuperscript{1}UNC Chapel Hill, \textsuperscript{2}UCSD, \qquad
Correspondence: \email{nurislam@cs.unc.edu}}
\maketitle              % typeset the header of the contribution
\begin{abstract}
Diffusion models, while trained for image generation, have emerged as powerful foundational feature extractors for downstream tasks. We find that off-the-shelf diffusion models, trained exclusively to generate natural RGB images,  can identify semantically meaningful correspondences in medical images. Building on this observation, we propose to leverage diffusion model features as a similarity measure to guide deformable image registration networks. We show that common intensity-based similarity losses often fail in challenging scenarios, such as when certain anatomies are visible in one image but absent in another, leading to anatomically inaccurate alignments. In contrast, our method identifies true semantic correspondences, aligning meaningful structures while disregarding those not present across images. We demonstrate superior performance of our approach on two tasks: multimodal 2D registration (DXA to X-Ray) and monomodal 3D registration (brain-extracted to non-brain-extracted MRI). Code: \href{https://github.com/uncbiag/dgir}{https://github.com/uncbiag/dgir}

%\keywords{Image Registration  \and Diffusion Models.}
\end{abstract}
\section{Introduction}

%Image registration is a fundamental technique in medical image analysis to estimate a spatial mapping between images. These images may come from different viewpoints, patients, time points, or imaging modalities. By ensuring that corresponding anatomies are accurately matched, image registration plays a crucial role in the accurate diagnosis, treatment planning, and monitoring of various medical conditions~\cite{fu2020deep}. 

%Deformable image registration enables accurate alignment of images with significant anatomical variations and can estimate complex transformations beyond simple parametric transformations such as rigid or affine transformations. Given a moving and a target images, non-parametric image registration estimates a dense deformation field between the image pair. This flexibility is particularly crucial in medical applications where natural variations in tissue shapes, patient movement, or organ changes over time need to be accurately accounted for~\cite{sotiras2013deformable}. 

Deep learning based deformable image registration~\cite{yang2017quicksilver,balakrishnan2019voxelmorph,fu2020deep} has shown to be useful in practice by delivering fast and accurate estimates of spatial correspondences (deformation fields) to align medical images. Deep networks are trained on populations of image pairs; typically by minimizing the sum of two losses: a similarity loss between the warped moving and the original fixed images and a regularization loss to encourage smoothness of the predicted deformation field. %Common similarity losses are based on local pixel intensities, such as Mean Squared Error (MSE), Localized Normalized Cross-Correlation (LNCC), Normalized Gradient Fields (NGF), and the Modality Independent Neighborhood Descriptor (MIND). The selection of a similarity measure depends on the specific characteristics of the dataset. For instance, MSE and NCC show good performance when registering images of the same modality, while NGF and MIND are popular for multimodal registration tasks. 

Conventional similarity losses (Mean Squared Error(MSE)~\cite{balakrishnan2019voxelmorph}, Localized Normalized Cross-Correlation (LNCC)~\cite{balakrishnan2019voxelmorph}, Normalized Gradient Fields (NGF)~\cite{haber2006intensity}, Modality Independent Neighborhood Descriptors (MIND)~\cite{heinrich2012mind}) assume identical anatomical structures in images and fail when anatomies are missing in one of the images~\cite{demir2024multimodal}. The "missing anatomy" scenario often appears in practice both in monomodal and multimodal registration. For example, to align a DXA (Dual Energy X-Ray) scan, which focuses on bone density and ignores soft tissues, to a standard X-Ray, which visualizes soft tissue in addition to bone, the model may stretch the bones in the DXA scan to fill the soft tissue space (see Fig. \ref{fig2dknee}).

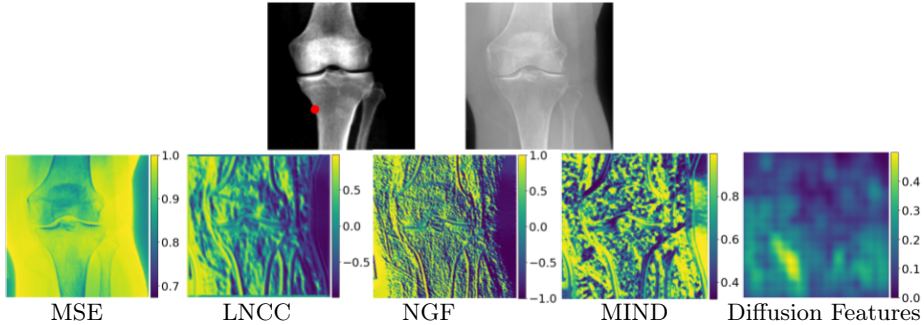
\begin{figure}[t]
\centering
\begin{minipage}{\linewidth}
\begin{picture}(100,109)
\put(0,6){\includegraphics[width=\linewidth]{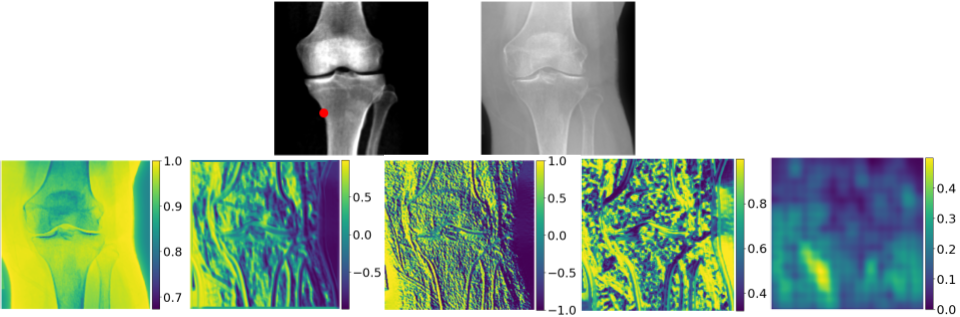}}

\put(17,0){MSE}
\put(82,0){LNCC}
\put(150,0){NGF}
\put(225,0){MIND}
\put(273,0){Diffusion Features}

\end{picture}
\end{minipage}
\caption{\textbf{Heatmaps.} First row: DXA scan (a red dot on the boundary between the bone and the background) and X-Ray. Second row: Heatmaps for different similarity measures indicating how close each point in the X-Ray is to the corresponding red point in the DXA scan. Correspondences for conventional pixel-based similarity measures are ambiguous as many  pixels show similar values, while diffusion features identify semantically meaningful correspondences. DXA
images reproduced by kind permission of the UK Biobank\textsuperscript{\textregistered}}
\label{f1}
\end{figure}

In monomodal registration, challenges arise when aligning pre- and post-operative images, or images at differing processing stages, such as magnetic resonance images (MRIs) of the head with and without brain extraction. Here, registration networks often attempt to compensate for the missing structures by expanding existing ones, as conventional similarity losses align contrasting edges, such as those between background and outermost anatomies, instead of desired internal alignment (see Fig.~\ref{f1}, \ref{fig2dknee}, \ref{fig3dbrain}). Therefore, for images with missing structures anatomy-aware losses with deep semantic understanding is required.

Diffusion models~\cite{ho2020denoising,rombach2022high} have revolutionized computer vision with high-fidelity image generation. Beyond achieving state-of-the-art image synthesis~\cite{dhariwal2021diffusion,rombach2022high}, the impressive generative abilities of these models suggest that they capture both low-level and high-level features, highlighting their potential as general representation learners~\cite{fuest2024diffusion}. These representations have been found useful in many downstream tasks such as for classification \cite{mukhopadhyay2024do,xiang2023denoising,chen2024deconstructing}, segmentation \cite{baranchuk2021label,tursynbek2023unsupervised,tian2024diffuse}, editing \cite{tumanyan2023plug,parmar2023zero,shi2024dragdiffusion}, depth and geometry estimation \cite{zhao2023unleashing,ke2024repurposing,lee2024exploiting}. 

\begin{figure}[t]
    \centering
    \begin{subfigure}[t]{0.4\textwidth}
        \centering
        \includegraphics[width=\linewidth, trim={0 3.4cm 0 0},clip]{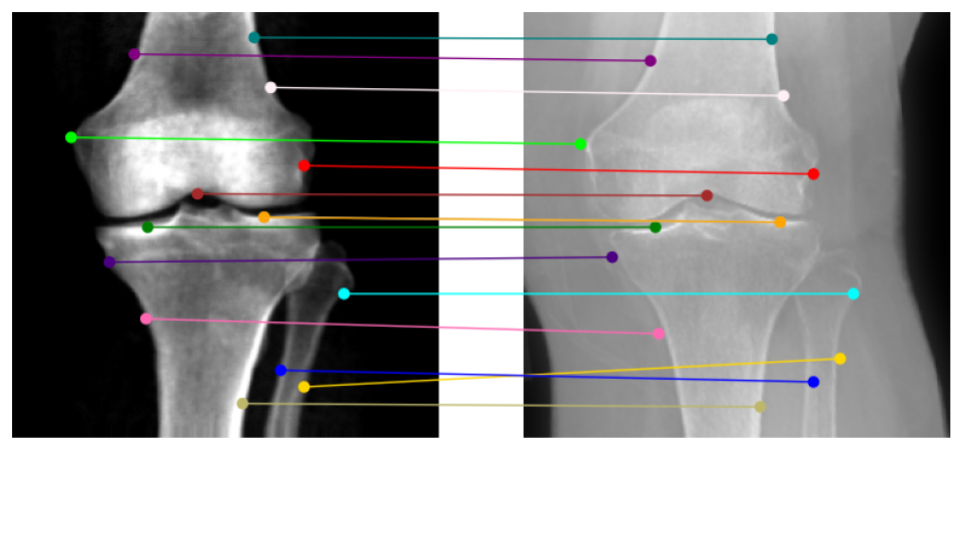}
        \caption{Diffusion Features.}
        \label{figcorresponda}
    \end{subfigure}%
    \hfill
    \begin{subfigure}[t]{0.4\textwidth}
        \centering
        \includegraphics[width=\linewidth, trim={0 3.4cm 0 0},clip]{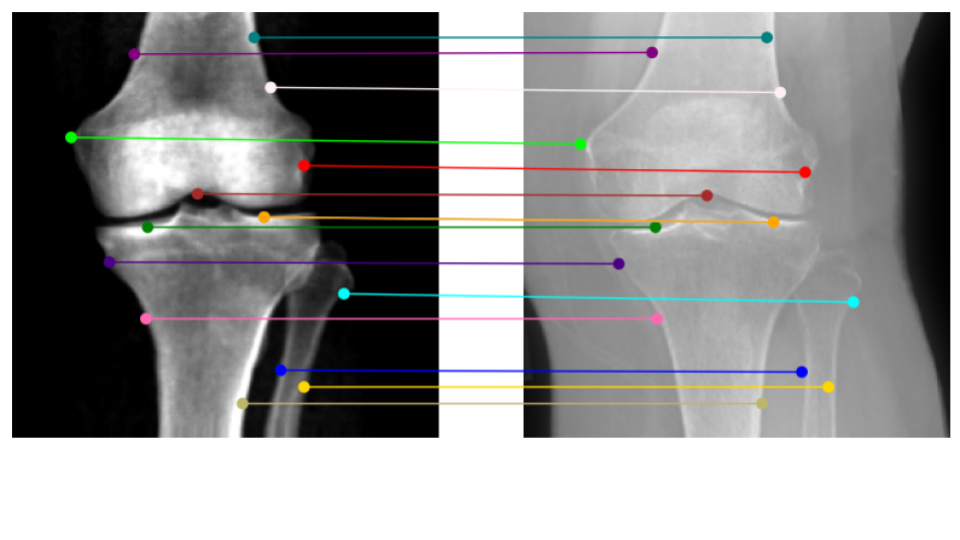}
        \caption{Diffusion Features + LNCC}
    \end{subfigure}
    \caption{\textbf{Correspondences.} Intermediate diffusion features, resized to image resolution, are vector-valued descriptors for each pixel. For each keypoint pixel in the left (DXA) the most similar (cosine similarity) pixel in the right (X-Ray) is connected with a line. (a) Diffusion features find complex correspondences in medical images with missing anatomies, although trained on RGB images. (b) LNCC on diffusion features further improves correspondences, considering surrounding pixels. DXA
    images reproduced by kind permission of the UK Biobank\textsuperscript{\textregistered}}
    \label{figcorrespond}
\end{figure}

Moreover, it has been shown that features from diffusion models, trained solely for image generation, have an emergent knowledge of semantic landmarks and correspondences even under challenging conditions such as occlusions and appearance variations~\cite{tang2023emergent,luo2024diffusion,meng2024not}. Intriguingly, this semantic correspondence ability extends to medical images (see Fig.~\ref{figcorresponda}), despite the significant domain gap and the absence of medical images in the training. Therefore, diffusion features may provide a useful signal for registration, especially in challenging scenarios (such as "missing anatomy"), where deep semantic understanding is necessary. %Equipped with this insight and the fact that a diffusion feature extractor (adding noise and propagating through a diffusion model) is a differentiable operator, we design a similarity loss that utilizes diffusion features to guide registration networks.

\textbf{Our contributions:}
\begin{itemize}
    \item We observe that diffusion models, trained to generate natural RGB images, also capture correspondences in medical images even in multimodal scenarios or where  not all anatomies are present in an image pair.
    \item We propose a registration similarity loss based on comparing diffusion features between a warped moving and a fixed image.
    \item We show superior performance on 2D registration of DXA to X-Ray and 3D registration of brain-extracted MRI to non-brain-extracted MRI.
\end{itemize}

\section{Related work}

\quad\;\;\textbf{Diffusion Models and Image Registration} have been studied together in the following ways. DiffuseMorph~\cite{kim2021diffusemorph} jointly trains diffusion and registration networks sequentially, passing noise predicted from the diffusion network as input to the registration network. FSDiffReg~\cite{qin2023fsdiffreg} extends this approach by adding intermediate diffusion features as input to the registration network. Diff-Def~\cite{starck2024diff} trains a diffusion model to generate deformation fields conditioned on specific parameters, guided by a registration network to optimize atlas deformation. DiffuseReg~\cite{zhuo2024diffusereg} combines diffusion and registration by iteratively denoising a deformation field from noise. While the concept of jointly training diffusion and registration networks is well explored, our work is the first work to explore off-the-shelf pre-trained diffusion models to define a similarity loss for registration. %\newline

%One exciting approach is to use diffusion to generate deformation fields for training a \emph{supervised} registration model. I remember reading this paper and quite liked it, but can't find it which will make it difficult to cite.

\textbf{Similarity losses for image registration} are well-studied for optimization-based~\cite{avants2011reproducible} and deep-learning approaches~\cite{hoopes2022learning,balakrishnan2019voxelmorph}. Common choices for monomodal registration are MSE and LNCC~\cite{balakrishnan2019voxelmorph}, while for multimodal registration - MIND~\cite{heinrich2012mind} and NGF~\cite{haber2006intensity}. These losses work well when images share anatomical structures, however they often fail when some anatomies are visible in one image and absent in another~\cite{demir2024multimodal}. Seg-Guided-MMReg~\cite{demir2024multimodal} guides multimodal registration with segmentaions as they are same across modalities. We find that images with missing anatomies not only share segmentation masks, but also have similar diffusion features. DeepSim~\cite{czolbe2021semantic} uses cosine similarity between features of a pre-trained autoencoder (or segmentation model) as a similarity loss. Unfortunately, it only works for monomodal scenarios with shared anatomical structures~\cite{czolbe2021semantic} and for each dataset, its own feature extractor needs to be trained, while diffusion models are training-free general feature extractors. Several works \cite{song2024general,kogl2024general} used non-learning optimization-based registration on foundational self-supervised features\cite{caron2021emerging,oquab2024dinov2}. We show how to incorporate knowledge from foundational diffusion models into the training of deep neural networks for registration.

\section{Method}

%\subsection{Preliminaries of diffusion models}

Diffusion models \cite{ho2020denoising} are composed of a forward and a reverse process. The forward process is a $T$-step noising of a clean image $\mathbf{x}_0\in\mathcal{R}^{H\times W\times C}$ with added Gaussian $\bm{\epsilon}\in\mathcal{R}^{H\times W\times C}$ at each step following a noise schedule $\{\beta_t\in\mathcal{R}^1\}_{t=1}^T$, slowly corrupting the image into noise $\mathbf{x}_T\in\mathcal{R}^{H\times W\times C}$. The noisy image at timestep $t$ is $\mathbf{x}_t=\sqrt{\overline{\alpha}_t}\mathbf{x}_0 + \sqrt{1-\overline{\alpha}_t}\bm{\epsilon}$, where $\overline{\alpha}_t = \prod_{i=1}^t(1-\beta_i)\in\mathcal{R}^1$. The $T$-step reverse process predicts the noise $\bm{\epsilon}$ at each timestep $t$ with a network $\mathbf{h}$: %At each timestep $t \in \llbracket 1, T \rrbracket$, the forward process is defined by a conditional distribution:
%\begin{equation}
%    q(\mathbf{x}_t|\mathbf{x}_{t-1}):=\mathcal{N}(\mathbf{x}_t; \sqrt{1-\beta_t}\mathbf{x}_{t-1}, \sqrt{\beta_t}\mathbf{I})\,,
%\end{equation}
%where $\{\beta_t\}_{t=1}^T$ is a pre-defined variance schedule that controls the rate of noise addition. With $\overline{\alpha}_t = \prod_{i=1}^t(1-\beta_i)$, the noisy image $\mathbf{x}_t$ can be expressed as \cite{ho2020denoising}:
%\begin{equation}
%    \mathbf{x}_t = \sqrt{\overline{\alpha}_t}\mathbf{x}_0 + \sqrt{1-\overline{\alpha}_t}\bm{\epsilon}\,, \qquad \bm{\epsilon}\sim\mathcal{N}(\mathbf{0},\mathbf{I})\,.
    %\label{eq2}
%\end{equation}

\begin{equation}
    \widehat{\bm{\epsilon}} = \mathbf{h}(\mathbf{x}_t, t)=\mathbf{h}(\sqrt{\overline{\alpha}_t}\mathbf{x}_0 + \sqrt{1-\overline{\alpha}_t}\bm{\epsilon}, t)\,, \qquad \bm{\epsilon}\sim\mathcal{N}(\mathbf{0},\mathbf{I})\,.
    \label{diffeq}
\end{equation}
The diffusion network $\mathbf{h}$ is trained to minimize the mean squared error (MSE) between the true and predicted noises $\|\bm{\epsilon} - \widehat{\bm{\epsilon}}\|^2$. Importantly, we do not train the diffusion model ourselves and use an off-the-shelf pre-trained diffusion model.

\subsection{Diffusion-Guided Image Registration}

Conventional deep learning methods train a registration network $\mathbf{f_\theta}$, with weights $\theta$, that takes moving image $\mathbf{A}$ and the fixed image $\mathbf{B}$ as inputs and outputs deformation map $\mathbf{\Phi}$, which warps $\mathbf{A}$ to the space of the fixed image, $\mathbf{A}\circ\mathbf{\Phi}$. Training involves minimizing a loss combining \textit{similarity} and \textit{regularization} terms:

\begin{equation}
    \min_{\theta}~ \mathcal{L}_{sim}(\mathbf{A}\circ\mathbf{\Phi}, \mathbf{B}) + \lambda\mathcal{L}_{reg}(\mathbf{\Phi})\,, \lambda>0\,.
\end{equation}

As noted earlier, conventional similarity losses focus on aligning strong edges, rather than desired internal structures. While this works when both images show the same anatomies, it often causes misalignment when some anatomies are missing in one image but present in the other. To address this, we propose Diffusion-Guided Image Registration (DGIR), leveraging features from pre-trained diffusion models, found to capture semantic correspondences, for image registration. 

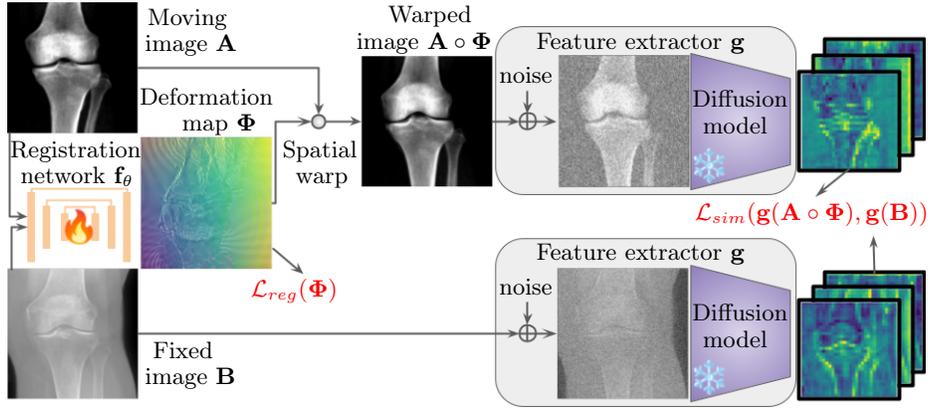
\begin{figure}[t]
\centering
\begin{minipage}{\linewidth}
\begin{picture}(200,154)
\put(0,0){\includegraphics[width=\linewidth]{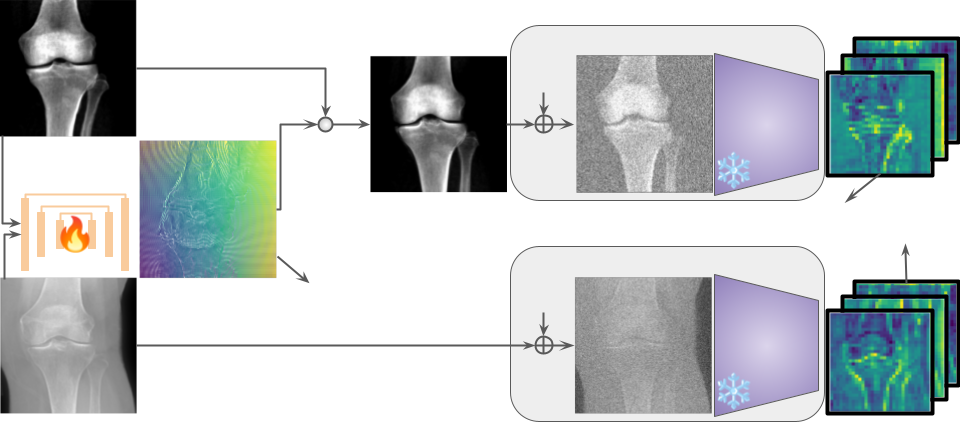}}
\put(53,145){Moving}
\put(52,135){image $\mathbf{A}$}
\put(55,19){Fixed}
\put(52,9){image $\mathbf{B}$}
\put(1,95){Registration}
\put(5,85){network $\mathbf{f_\theta}$}
\put(50,115){Deformation}
\put(66,105){map $\mathbf{\Phi}$}
\put(104, 95){Spatial}
\put(109, 85){warp}
\put(144,146){Warped}
\put(132,136){image $\mathbf{A}\circ\mathbf{\Phi}$}
\put(260,71){\textcolor{red}{$\mathcal{L}_{sim}(\mathbf{g}(\mathbf{A}\circ\mathbf{\Phi}), \mathbf{g}(\mathbf{B}))$}}
\put(92,42){\textcolor{red}{$\mathcal{L}_{reg}(\mathbf{\Phi})$}}
\put(259,114){Diffusion}
\put(263,104){model}
\put(200,136){Feature extractor $\mathbf{g}$}
\put(186,123){noise}
\put(259,33){Diffusion}
\put(263,23){model}
\put(200,56){Feature extractor $\mathbf{g}$}
\put(186,43){noise}
\end{picture}
\end{minipage}
\caption{\textbf{Overview of the method.} The registration network $\mathbf{f}_\theta$ is trained with a similarity loss that encourages diffusion features $\mathbf{g}(\mathbf{A}\circ\mathbf{\Phi})$ of the warped image  $\mathbf{A}\circ\mathbf{\Phi}$ to be similar to the diffusion features $\mathbf{g}(\mathbf{B})$ of the fixed image $\mathbf{B}.$ A regularization loss is applied to the deformation map $\mathbf{\Phi}$ for spatial smoothness. DXA images reproduced by kind permission of the UK Biobank\textsuperscript{\textregistered}}
\label{foverview}
\end{figure}

Fig.~\ref{foverview} shows an overview of our framework. The input (moving image $\mathbf{A}$ and fixed image $\mathbf{B}$) and output (deformation map $\mathbf{\Phi}$) of the registration network $\mathbf{f}_\theta$ are the same as for typical registration networks. DGIR additionally passes warped and fixed images through a diffusion-based feature extractor $\mathbf{g}$:

\begin{equation}
    \mathbf{g}(\mathbf{x}) = \overline{\mathbf{h}}_n(\sqrt{\overline{\alpha}_t}\mathbf{x} + \sqrt{1-\overline{\alpha}_t}\bm{\epsilon}, t)\, ,
\end{equation}
where $\overline{\mathbf{h}}_n$ is $n$-th block output of the pre-trained diffusion model $\mathbf{h}$ from Eq.~\eqref{diffeq}. %The noise $\bm{\epsilon}\sim\mathcal{N}(\mathbf{0},\mathbf{I})$ and the cumulative product $\overline{\alpha}_t = \prod_{i=1}^t(1-\beta_i)$ are based on the diffusion schedule $\{\beta_t\}_{t=1}^T$.

Instead of defining a similarity loss directly between the warped and the fixed images, we compute it on the diffusion features to promote deep semantic alignment. Note that these diffusion features are computed on the \emph{warped} moving image and the fixed image separately. I.e., for a perfect image alignment these features would by construction be identical. We use the $1-LNCC$ loss on the diffusion features as our similarity loss and spatial gradient penalization~\cite{balakrishnan2019voxelmorph} of the displacement field $\|\nabla(\mathbf{\Phi}-\textbf{id})\|^2_F$ (where $\textbf{id}$ is the identity map) as regularization:

\begin{equation}
    \mathcal{L} = \mathcal{L}_{sim}(\mathbf{g}(\mathbf{A}\circ\mathbf{\Phi}), \mathbf{g}(\mathbf{B}))+\lambda\mathcal{L}_{reg}(\mathbf{\Phi}) = 1 - LNCC(\mathbf{g}(\mathbf{A}\circ\mathbf{\Phi}), \mathbf{g}(\mathbf{B})) +\lambda \|\nabla(\mathbf{\Phi}-\textbf{id})\|^2_F\,,
    \label{ourloss}
\end{equation}

\label{dgir}

\subsection{Datasets and Implementation Details}
\label{details}

\textbf{Registration}. For all registration networks, we use a multi-step U-Net~\cite{greer2021icon} and train with Adam and learning rate $1e-4$. Models are implemented using PyTorch and trained on one Nvidia A6000 GPU. We set  $\lambda=1$ for all experiments.

\noindent\textbf{Diffusion Models}. We use a publicly available unconditional diffusion model from~\cite{dhariwal2021diffusion} pre-trained on $256\times256\times3$ RGB images from ImageNet~\cite{deng2009imagenet}. 
This model is a U-Net \cite{ronneberger2015u} with 37 blocks (18 Encoder, 1 Middle, and 18 Decoder blocks). 

\noindent\textbf{2D experiments}. We use knee DXA scans from the UK Biobank~\cite{sudlow2015uk} and X-Rays from The Osteoarthritis Initiative (OAI)~\cite{nevitt2006osteoarthritis}. %This dataset is multimodal and has missing anatomies as soft tissue is not visible in the DXA scans but visible in the X-Rays. 
Following~\cite{demir2024multimodal}, we resize the images to $256\times256$ and affine aligned them to a random reference image using the affine layers of \cite{greer2023inverse}. We use $98$ DXA scans and $758$ X-Rays as a training set and $25$ DXA scans and $100$ X-Rays as a test set ($2500$ test pairs) and manual segmentations of the 3 major knee bones (femur, tibia, and fibula) for evaluation. %We repeat images 3 times along the channel dimension to make them compatible with expected input for the diffusion model. %If it's not written we extract output activations at block $n=$, set regularization parameter $\lambda=$, add diffusion noise of strength $t=$. 

\noindent\textbf{3D experiments}. Although the diffusion model is pre-trained on 2D RGB images, it provides useful signals for \emph{volumetric} 3D registration. For that, we randomly select $N$ coronal, or sagittal, or axial slices for both fixed and warped 3D images. Then, we extract diffusion features for these slices and apply the LNCC loss on them. Throughout the training, this process will cover the entire volume thereby using a 2D feature extractor for 3D registration.  We use 3D brain MRIs from Neurite-OASIS~\cite{marcus2007open,hoopes2022learning}, which consists of $414$ 3D images affinely alligned to  FreeSurfer Talairach atlas \cite{fischl2012freesurfer} and center-cropped to size $160\times192\times224$. The dataset comes in two forms: raw (non-brain-extracted) MRIs and brain-extracted/skull-stripped MRIs. We use $300$ images as a training set, and $114$ images as a test set ($114\times113=12882$ test pairs) and 4 brain region segmentations (Cortex, White Matter, Gray Matter, CSF) for evaluation.

%\vspace{-0.4cm}

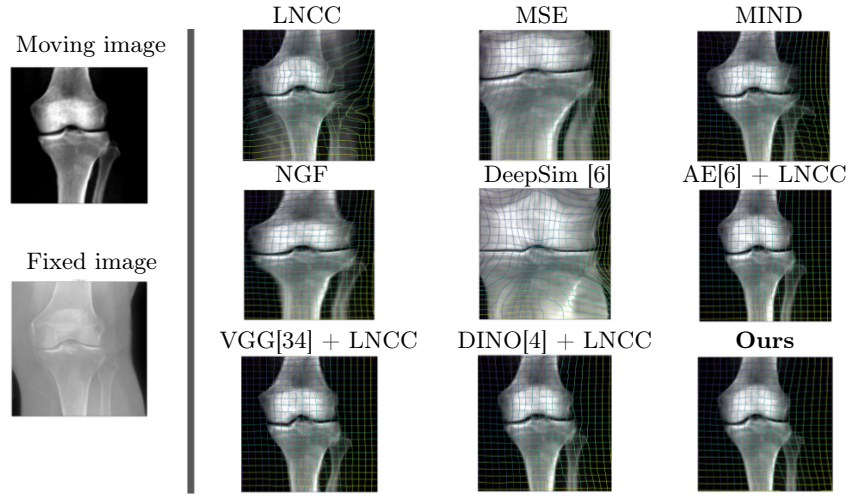
\begin{figure}[t]
\centering
\begin{minipage}{\linewidth}
\begin{picture}(200,180)
\put(10,0){\includegraphics[width=0.9\linewidth]{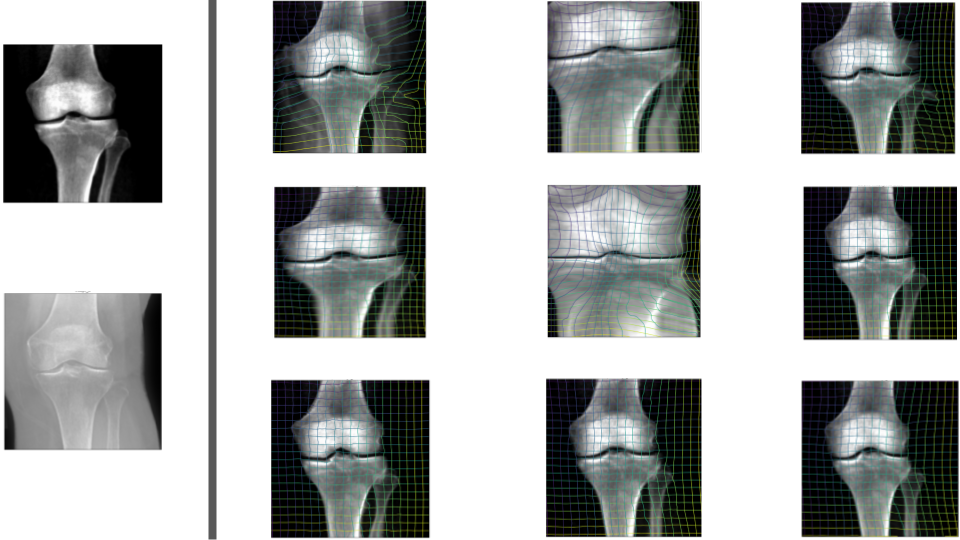}}

\put(13,167){Moving image}
\put(17,85){Fixed image}
\put(110,178){LNCC}
\put(202,178){MSE}
\put(285,178){MIND}

\put(111,118){NGF}
\put(190,118){DeepSim \cite{czolbe2021semantic}}
\put(265,118){AE\cite{czolbe2021semantic} + LNCC}

\put(90,55){VGG\cite{simonyan2014very} + LNCC}
\put(180,55){DINO\cite{caron2021emerging} + LNCC}
\put(285,55){\textbf{Ours}}

\end{picture}
\end{minipage}
\caption{\textbf{2D results.} Pixel-based similarity losses fail to capture true correspondences in the "missing anatomy" scenario. Images on the right show the warped moving image which should ideally resemble the fixed image. Our method (Diffusion Features + LNCC) performs the best, as quantitatively shown in Tab.~\ref{table_knee}. DXA images reproduced by kind permission of the UK Biobank\textsuperscript{\textregistered}}
\label{fig2dknee}
\end{figure}

\section{Multimodal (DXA$\rightarrow$X-Ray) 2D Knee Registration}

\label{2d_knee}

%The baseline similarity losses we compare to are LNCC~\cite{balakrishnan2019voxelmorph}, MIND~\cite{heinrich2012mind}, DeepSim~\cite{czolbe2021semantic}, MSE~\cite{balakrishnan2019voxelmorph}, and NGF~\cite{haber2006intensity}. 

We show registration results in Fig.~\ref{fig2dknee} and report Dice scores of bone segmentations and percentage of pixels with negative Jacobian for the registration map, $\Phi$, in Tab.~\ref{table_knee}. Conventional methods fail in finding appropriate semantic correspondences in the "missing anatomy" setting. They stretch the bones and fill up the space of soft tissue, which is visible in X-Rays and is missing in DXA scans. 

Although other feature extractors (Autoencoder~\cite{czolbe2021semantic}, VGG~\cite{simonyan2014very}, DINO~\cite{caron2021emerging}) do not have this issue, the features from off-the-shelf pretrained diffusion models show the best performance (Tab.~\ref{table_knee}). The combination of LNCC with diffusion features outperforms all other baselines and even the existing state-of-the-art method Seg-Guided-MMReg~\cite{demir2024multimodal}, which was trained with segmentations. %This shows that our approach is a strong unsupervised method that achieves state-of-the-art registration performance in the challenging scenario of registering multimodal images with missing anatomies.

%The combination of using LNCC similarity on deep features indeed helps to find anatomical correspondences. Compared to other feature extractors, training registration networks using feaures from a diffusion model  performs the best (see Table \ref{table_knee}). 

%To emphasize that pre-trained diffusion-based feature extractors are indeed a better choice for registration, we also replace them with other feature extractors for comparison, all using the same LNCC on the feature space. Namely, we use VGG~\cite{simonyan2014very}, DINO~\cite{caron2021emerging}, and a dataset-specific autoencoder feature extractor, which was trained to reconstruct both types of images (DeepSim~\cite{czolbe2021semantic} with LNCC) . %This baseline is similar to DeepSim, except instead of cosine similairty of deep features we use LNCC on the deep features of the autoencoder.

%The baseline similarity losses we compare to are LNCC~\cite{balakrishnan2019voxelmorph}, MIND~\cite{heinrich2012mind}, DeepSim~\cite{czolbe2021semantic}, MSE~\cite{balakrishnan2019voxelmorph}, and NGF~\cite{haber2006intensity}. Figure \ref{fig2dknee} shows qualitative results for 2D knee registration experiments. We observe that conventional methods indeed fail in finding appropriate semantically meaningful correspondences even between common anatomical structures when one of the images contains anatomical structures that are not visualized in the other image. 

\begin{figure}[t]
    \centering
    \begin{subfigure}[t]{0.45\textwidth}
        \centering
        \includegraphics[width=\linewidth, trim={0 0.3cm 0 0},clip]{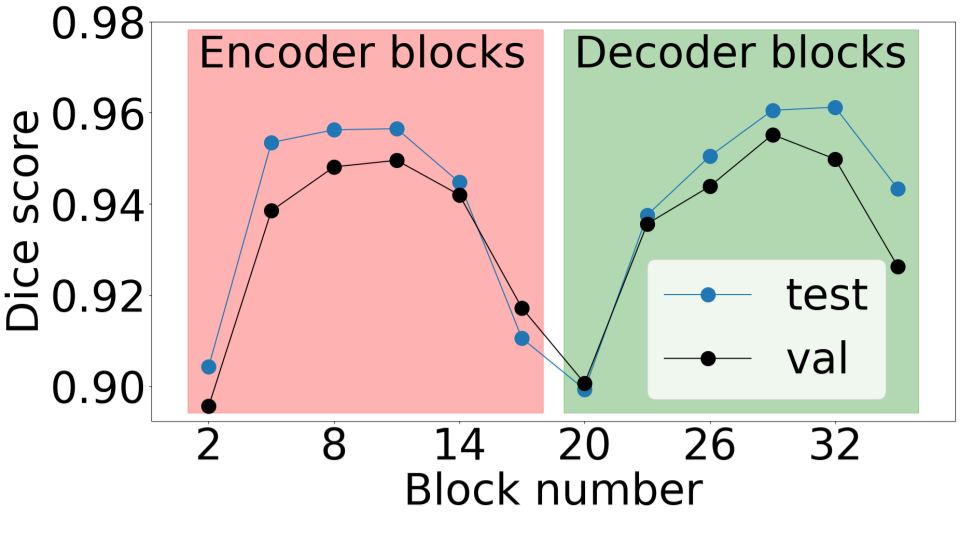}
        \caption{Mid-resolution features are the best to guide registration. Decoder features are slightly better than encoder ones.}
        \label{layer}
    \end{subfigure}%
    ~ 
    \begin{subfigure}[t]{0.45\textwidth}
        \centering
        \includegraphics[width=\linewidth]{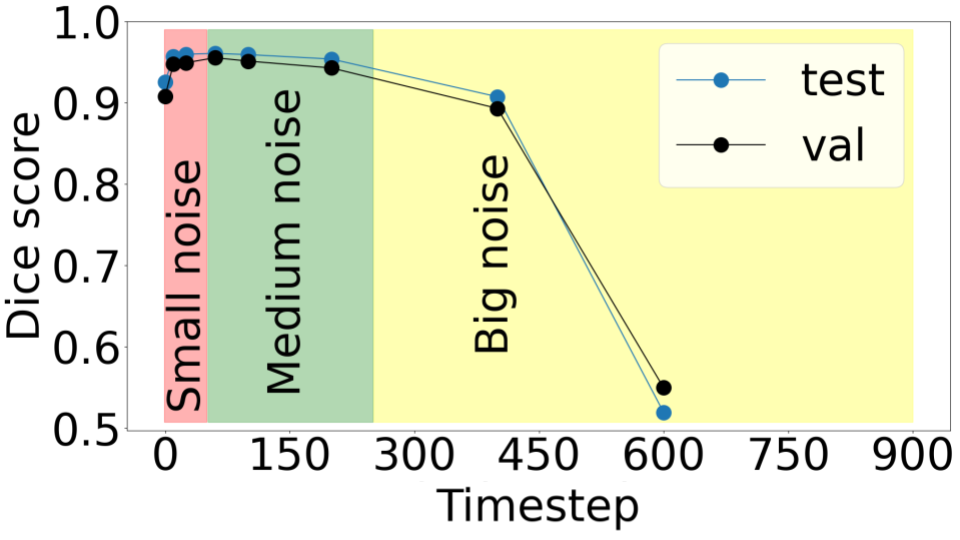}
        \caption{Adding small-to-medium noise before passing an image to a pretrained diffusion model works best.} 
        \label{timestep}
    \end{subfigure}
    \caption{\textbf{Block number and Noise timestep ablations on 2D knee dataset.}}
    \label{figablations}
\end{figure}

The off-the-shelf pre-trained diffusion model from~\cite{dhariwal2021diffusion} uses a U-Net with $37$ blocks and $T=1000$ noise levels (timesteps). We ablated outputs of which block and what timestep  to use to guide a registration network in Fig.~\ref{figablations}. In Fig.~\ref{layer} we fix the timestep at $t=60$ and measure the average Dice scores for the networks guided with different diffusion blocks. Mid-resolution blocks (8-11 and 29-32) perform better compared to others and decoder blocks are slightly better than encoder blocks. In Fig.~\ref{timestep}, we fix the block number $n=29$ and measure the average Dice scores for the networks guided with different noise timesteps. The best performance is observed when we add small-to-medium ($t\in[10;150]$) noise. Test dice scores for our method reported in Tab.~\ref{table_knee} are based on the model with best parameters (timestep $t=50$, block $n=29$) on small validation set.%It should be mentioned that adding slight noise ($t\in[10;150]$) is better than using a clean image to the diffusion feature extractor. Performance significantly drops at large  noise levels ($t>250$). This is expected as large noise destroys most of the useful image information.

\begin{table}[t]
    \centering
    \caption{\textbf{Quantitative 2D knee registration results.} Dice scores of multimodal registration with missing anatomy between DXA scans and X-Rays.}
    \begin{tabular}{l@{\hskip 0.18in}c@{\hskip 0.18in}c@{\hskip 0.18in}c@{\hskip 0.18in}c@{\hskip 0.18in}c@{\hskip 0.18in}c}
    \toprule
    Method & Femur  & Tibia  & Fibula & Average  & \%|J| \\
         \toprule
         LNCC \cite{balakrishnan2019voxelmorph} & 0.8209 & 0.8300 & 0.7323 &  0.7944 & 0.00\% \\
         MSE \cite{balakrishnan2019voxelmorph} & 0.6267 & 0.6938 & 0.4733 & 0.5980 & 0.00\% \\
         MIND \cite{heinrich2012mind} & 0.9267 & 0.9407 & 0.8236& 0.8970  & 0.00\%\\
         NGF \cite{haber2006intensity}  & 0.8162 & 0.8350 & 0.4724 & 0.7079 & 0.00\%\\
         DeepSim \cite{czolbe2021semantic} & 0.6808 & 0.6452 & 0.0868 & 0.4709 & 0.00\%\\
         AE \cite{czolbe2021semantic} + LNCC & 0.9423 & 0.9490 & 0.7582 & 0.8832 & 0.00\%\\
         VGG \cite{simonyan2014very} + LNCC & 0.9446 & 0.9368 &  0.7207 & 0.8674 & 0.00\% \\
         DINO \cite{caron2021emerging} + LNCC & 0.9621 & 0.9584 & 0.8603 & 0.9269 & 0.00\% \\
         \textbf{Diffusion + LNCC(Ours)} & \textbf{0.9804} & \textbf{0.9739} & \textbf{0.9291} & \textbf{0.9611} & 0.00\% \\
         \midrule
         Seg-Guided-MMReg \cite{demir2024multimodal} & 0.9701 & 0.9684 & 0.8882 & 0.9422 & 0.00\%\\
         \bottomrule
    \end{tabular}
    \label{table_knee}
\end{table}

%Additionally, we report Average Dice with Instance Optimization (IO), where for each test pair we finetune the registration network for $50$ more iterations. These results indeed verify that our loss finds correct correspondences and optimize deformation field in a right way

\section{3D brain registration}

%We fix the timestep at $t=50$ and use block $n=11$, which provides comparable registration performance to the corresponding decoder block (see Figure \ref{layer}), while significantly saving time and memory, essential for 3D registrations.

We leverage representations from 2D diffusion to guide 3D registration with missing anatomy (for brain MRIs: e.g., skull, neck). We fix timestep $t=50$ and $n=11$ block, which is comparable to the corresponding decoder block (see Fig.~\ref{layer}), while significantly saves time and memory, essential for 3D registrations.

Our goal is not to push the State-Of-The-Art (SOTA) in specific registration but rather to investigate the broader utility of our loss function in a general context. We compare to two networks trained with 3D LNCC~\cite{balakrishnan2019voxelmorph}: Brain-Extracted moving to Non-Brain-Extracted fixed images (BE$\rightarrow$NBE) and Brain-Extracted moving to Brain-Extracted fixed images (BE$\rightarrow$BE) We also compare to foundational registration method, UniGradICON~\cite{tian2024unigradicon}, which was trained on diverse and extensive collection of medical images with LNCC similairty. Our network is  Brain-Extracted moving to Non-Brain-Extracted fixed images (BE$\rightarrow$NBE) trained with diffusion guidance (see Eq. \eqref{diffeq} and Fig.~\ref{foverview} in Section~\ref{dgir}). %We report average Dice score on $4$ brain region segmentations in Tab.~\ref{table_brain} and show qualitative results in Fig.~\ref{fig3dbrain}. 

\begin{figure}[t]
\centering
\begin{minipage}{\linewidth}
\begin{picture}(200,145)
\put(10,0){\includegraphics[width=\linewidth, trim={0 6.3cm 0 0},clip]{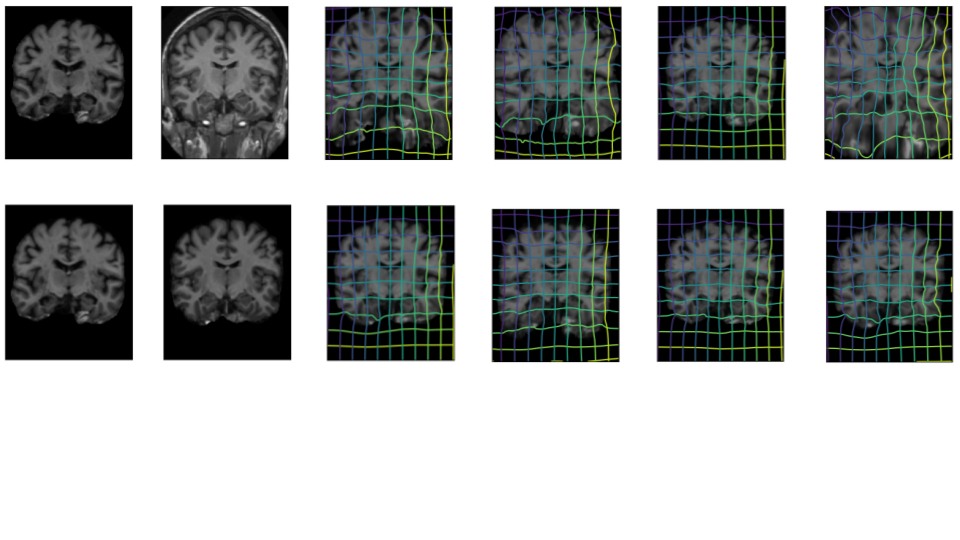}}

\put(20,140){Moving}
\put(20,130){Image}
\put(80,140){Fixed}
\put(80,130){image}
\put(130,140){LNCC \cite{balakrishnan2019voxelmorph}}
\put(120,130){Tr.: BE$\rightarrow$BE}
\put(195,140){LNCC \cite{balakrishnan2019voxelmorph}}
\put(180,130){Tr.: BE$\rightarrow$NBE}
\put(257,140){\textbf{Ours}}
\put(244,130){Tr.: BE$\rightarrow$NBE}

\put(299,140){UniGradICON}
\put(321, 130){\cite{tian2024unigradicon}}

\put(0, 0){\rotatebox{90}{Test: BE$\rightarrow$BE}}

\put(0, 65){\rotatebox{90}{Test: BE$\rightarrow$NBE}}

\end{picture}
\end{minipage}
\caption{\textbf{3D registration.} Pixel-based similarity (LNCC) methods andfoundational UniGradICON model fail to capture true correspondences in the "missing anatomy" test scenario (BE$\rightarrow$NBE), stretching the brain to the space of skull and neck. Our method reliably performs desired alignment in both scenarios.}
\label{fig3dbrain}
\end{figure}
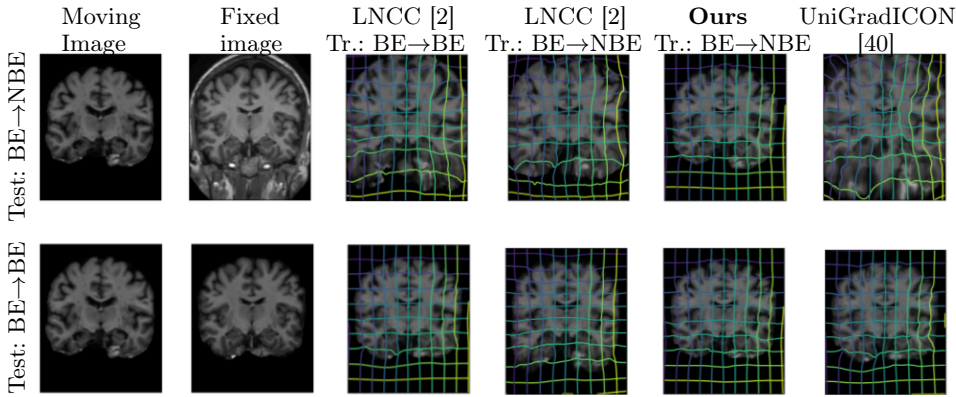

As we see in Tab.~\ref{table_brain} and Fig.~\ref{fig3dbrain}, both LNCC~\cite{balakrishnan2019voxelmorph} networks and foundational UniGradICON~\cite{tian2024unigradicon} fail to capture semantically meaningful correspondences and rely on edges with strong contrast, stretching the gray matter to the skull and neck region, while our method learns semantically meaningful alignment. It significantly outperforms other methods in the test BE$\rightarrow$NBE scenario and performs on par in the standard BE$\rightarrow$BE test scenario with foundational UniGradICON and an LNCC network that was specifically trained for this scenario. 

\begin{table}[t]
    \centering
    \caption{\textbf{3D results.} Our approach performs on par with the LNCC loss when moving and fixed images share anatomies and foundationl UniGradICON, while it  performs significantly better when one of the images has missing anatomy.}
    \begin{tabular}{lc@{\hskip 0.06in}c@{\hskip 0.06in}c@{\hskip 0.06in}c@{\hskip 0.06in}c}
    \toprule
    \quad\;\quad Method & \begin{tabular}{@{}c@{}} Dice Score \\ BE$\rightarrow$NBE test\end{tabular} & \begin{tabular}{@{}c@{}} \%|J| \\ BE$\rightarrow$NBE test\end{tabular}& \begin{tabular}{@{}c@{}} Dice Score \\ BE$\rightarrow$BE test\end{tabular} & \begin{tabular}{@{}c@{}} \%|J| \\ BE$\rightarrow$BE test\end{tabular} \\
         \midrule
         LNCC~\cite{balakrishnan2019voxelmorph} (Tr.:BE$\rightarrow$BE)
          & 0.6325 & 1.411\% & 0.8156  & 0.020\% \\
         LNCC~\cite{balakrishnan2019voxelmorph} (Tr.:BE$\rightarrow$NBE)& 0.5841 & 0.806\% & 0.6983 & 0.073\% \\
         \textbf{Ours (Tr.:BE$\rightarrow$NBE)}& 0.8111 & 0.633\% & 0.8079 & 0.268\%\\
         \midrule
         UniGradICON \cite{tian2024unigradicon} & 0.5326 & 0.002\% & 0.8380 & 0.001\%\\
         \bottomrule
    \end{tabular}
    \label{table_brain}
\end{table}

\section{Conclusion}

In this work, we presented Diffusion-Guided Image Registration (DGIR), a method inspired by the observation that pre-trained diffusion models trained solely for image generation have the emergent capability of finding semantic correspondences even in medical images. We show that diffusion-guided image registration is useful in challenging scenarios of missing anatomy (monomodal and multimodal), where deep semantic understanding is required to find same anatomies and perform accurate registration. A potential future work might be to train a large-scale volumetric diffusion model on multiple diverse datasets and see what emergent properties these models have. \\

\noindent\textbf{Acknoledgements:} This research was, in part, funded by the National Institutes of Health (NIH) under awards 1R01AR082684 and 1R01HL149877. The views and conclusions contained in this document are those of the authors and should not be interpreted as representing official policies, either expressed or implied, of the NIH. This research has been conducted using the UK Biobank Resource under Application Number 22783.

\bibliographystyle{splncs04}
\bibliography{main}

\end{document}